\pdfoutput=1

\documentclass[11pt]{article}

\usepackage[final]{acl} 

\usepackage{times}
\usepackage{latexsym}

\usepackage[T1]{fontenc}

\usepackage[utf8]{inputenc}

\usepackage{microtype}

\usepackage{inconsolata}

\usepackage{graphicx}

\usepackage{amsmath}
\usepackage{subcaption}
\title{CausalMamba: Interpretable State Space
Modeling for Temporal Rumor Causality}

\author{Xiaotong Zhan \\
  University of California, Berkeley \\
  Department of Statistics \\
  \texttt{Xiaotong\_zhan@berkeley.edu} \\\And
  Xi Cheng \\
  University of California, Berkeley \\
  Department of Statistics \\
  \texttt{xi\_cheng1017@berkeley.edu} \\}

\begin{document}
\maketitle
\begin{abstract}
Rumor detection on social media remains a challenging task due to the complex propagation dynamics and the limited interpretability of existing models. While recent neural architectures capture content and structural features, they often fail to reveal the underlying causal mechanisms of misinformation spread. We propose CausalMamba, a novel framework that integrates Mamba-based sequence modeling, graph convolutional networks (GCNs), and differentiable causal discovery via NOTEARS. CausalMamba learns joint representations of temporal tweet sequences and reply structures, while uncovering latent causal graphs to identify influential nodes within each propagation chain. Experiments on the Twitter15 dataset show that our model achieves competitive classification performance compared to strong baselines, and uniquely enables counterfactual intervention analysis. Qualitative results demonstrate that removing top-ranked causal nodes significantly alters graph connectivity, offering interpretable insights into rumor dynamics. Our framework provides a unified approach for rumor classification and influence analysis, paving the way for more explainable and actionable misinformation detection systems.
\end{abstract}

\section{Introduction}

The rapid spread of misinformation on social media platforms has raised growing concerns about its societal impact, from public health disinformation to political rumors. Automated rumor detection has thus become a critical task in natural language processing (NLP) and computational social science. While recent models have achieved promising accuracy using content-based features and neural network classifiers, they often operate as black boxes—lacking transparency and interpretability regarding why a piece of information is considered a rumor.

This lack of interpretability presents a serious limitation, particularly when decisions involve moderation or user-facing feedback. More importantly, traditional models do not support fine-grained causal analysis of how information spreads, nor do they identify which specific messages or users drive the diffusion process. Without such insight, proactive interventions remain out of reach.

Recent advances in sequence modeling and graph learning have introduced new opportunities for capturing the dynamics of rumor propagation. Mamba~\citep{gu2023mamba}, a selective state space model, has demonstrated impressive performance on long sequence tasks with linear efficiency, while graph neural networks (GNNs) can encode structural reply trees. However, few existing approaches combine these strengths with causal discovery—the ability to uncover latent influence structures from observed behavior.

In this work, we propose \textbf{CausalMamba}, a unified framework that integrates sequence modeling, graph representation learning, and differentiable causal discovery. Our model consists of three components: (1) a Mamba encoder to capture sequential content dependencies within a rumor cascade; (2) a GCN encoder to model the reply tree structure; and (3) a NOTEARS-based causal graph learner to extract directed acyclic graphs (DAGs) over tweets for influence attribution. The model is trained with a joint loss that combines classification and causal sparsity objectives.

Experiments on the Twitter15 dataset show that CausalMamba achieves competitive performance compared to strong baselines such as BiLSTM-CNN and Transformer variants. Furthermore, the learned causal graphs support intervention simulation—removing top-ranked nodes identified via PageRank significantly disrupts graph connectivity and alters predicted outcomes.

\textbf{Our main contributions} address three research questions (RQs): 

RQ1: Can we improve rumor classification by jointly modeling content, structure, and time? 

RQ2: Can we discover interpretable causal structures from rumor propagation chains? 

RQ3: Can we simulate node-level interventions to identify key drivers of misinformation? 

To address these questions, we propose CausalMamba, a unified framework that integrates Mamba encoders, graph convolutional networks, and differentiable causal discovery. Our model offers both strong predictive performance and interpret-ability, enabling actionable insights for understanding and mitigating the spread of misinformation.

\section{Related Work}

\subsection{Structured Sequence Modeling for Long-Range Dependencies}

Transformer-based architectures have achieved remarkable success across NLP tasks, but they exhibit limitations in modeling long-range compositional reasoning due to their quadratic attention complexity and limited memory capacity \citep{dziri2023faith, peng2024limitations}. To address this, Structured State Space Models (SSMs) have emerged as a compelling alternative. The S4 model \citep{gu2022s4} and its variants like DSS \citep{gupta2022dss} and S5 \citep{lu2023s5} improve efficiency by leveraging state-space parameterization for sequence modeling. The recent Mamba model \citep{gu2023mamba} introduces selective updates based on input content, outperforming Transformers on various long-context benchmarks. While promising, applications of SSMs in NLP remain limited, and their integration into rumor detection tasks is largely unexplored.

\subsection{Causal Discovery and Representation Learning}

Modern causal analysis frameworks aim to uncover underlying generative mechanisms beyond statistical correlation. NOTEARS \citep{zheng_dags_2018} reformulates causal graph discovery as a continuous optimization problem, enabling scalable DAG learning from observational data. Follow-up methods have extended this to nonlinear and time-series settings \citep{yu_dag-gnn_2019}, though most are applied outside the NLP domain. Within NLP, recent studies have explored the use of causal inference for tasks like debiasing and counterfactual reasoning \citep{feder_causal_2022, li_causal_2023}. However, learning causal graphs over textual propagation patterns remains an open challenge.

\subsection{Causality-Informed Rumor Detection}

Rumor detection has traditionally been framed as a supervised classification task using text features and propagation signals \citep{wu_misinformation_2019}. Recent efforts incorporate causal reasoning to enhance interpretability. For instance, \citet{cheng_causal_2021} mitigate confounding bias in user-news interactions, while \citet{hu2022causal} leverage image-text causality in multimodal misinformation detection. These works reveal the potential of causal modeling in social media analysis, yet they do not jointly consider graph-based structure and temporal sequences.

\subsection{Graph Neural Networks for Propagation Structure Modeling}

Graph neural networks (GNNs) have been widely used to capture structural patterns in rumor propagation. Among them, Graph Convolutional Networks (GCNs) have shown strong capabilities in encoding relational signals from propagation trees or retweet networks. For instance, \citet{bian_rumor_2020} propose Bi-GCN, which jointly models top-down and bottom-up information flows in a rumor tree to capture both propagation and dispersion patterns. Their approach demonstrates that leveraging graph directionality significantly improves early rumor detection performance. Similarly, \citet{dong2019multiple} address the problem of identifying multiple rumor sources using a GCN-based framework, showing that spectral graph convolutions can effectively capture multi-hop dependencies without prior assumptions on the underlying diffusion model.

Despite these successes, existing GCN-based models typically assume static propagation graphs and treat information flow as correlational. They lack mechanisms to distinguish genuine causal influence from incidental structural proximity, which can lead to overfitting or misinterpretation of node importance. Moreover, such models often do not integrate temporal or sequential content signals during learning.

Our work addresses these limitations by integrating a Mamba-based state space encoder for temporal textual modeling, a GCN for structural encoding, and a differentiable causal discovery module based on NOTEARS. This unified framework allows us to jointly learn semantic, structural, and causal representations for rumor detection and intervention simulation.

\subsection{Research Gaps}

While previous research has separately explored state space modeling, graph neural networks, and causal discovery, few works combine these strengths in an end-to-end manner. Our work bridges this gap by integrating Mamba-based encoding, GCN structural learning, and NOTEARS style causal graph discovery into a unified framework for interpretable rumor detection.

\section{Dataset}

We conduct all experiments on the \textbf{Twitter15} dataset, a widely-used benchmark for rumor detection introduced by \citet{ma2017detect}. Twitter15 contains 1,490 events, where each event begins with a source tweet and is followed by a cascade of replies and retweets forming a propagation tree. Each event is annotated with one of four veracity labels: \textit{True}, \textit{False}, \textit{Unverified}, or \textit{Non-rumor}. Although Twitter16 is often used as a complementary dataset, our current implementation focuses exclusively on Twitter15.

\noindent\textbf{Graph Construction.} For each event, we parse the propagation tree and build a directed graph where nodes represent tweets and edges reflect reply or retweet relationships. Nodes are sorted by actual propagation time to ensure temporal consistency, which is critical for downstream causal modeling. Redundant self-loops (e.g., $(0,0)$) are removed during preprocessing.

\noindent\textbf{Feature Engineering.} Each node is encoded using three feature types:
\begin{itemize}
    \item \textbf{Textual:} 768-dimensional BERT \texttt{[CLS]} embeddings (from \texttt{bert-base-uncased}), with input truncated to 128 tokens.
    \item \textbf{Temporal:} Log-transformed delay from the source tweet, using $\log(1 + \Delta t)$.
    \item \textbf{User:} 64-dimensional embedding from hashed user IDs, with salted hashing to mitigate collisions.
\end{itemize}
The final node representation is a concatenation of all features, yielding an 833-dimensional input.

\noindent\textbf{Cleaning and Splitting.} Events with fewer than two tweets are excluded. We retain 1,490 valid graphs with no missing values. The dataset is split into training (70\%), validation (15\%), and test (15\%) subsets with balanced label distribution. Table~\ref{tab:dataset_stats} summarizes the dataset.

\begin{table}[!htbp]
\centering
\begin{tabular}{l c}
\hline
\textbf{Metric} & \textbf{Twitter15} \\
\hline
Events & 1,490 \\
Avg. Nodes / Event & 408 \\
Avg. Edges / Event & 1,216 \\
Missing Values & 0 \\
\hline
\end{tabular}
\caption{Statistics of Twitter15 after preprocessing.}
\label{tab:dataset_stats}
\end{table}

Table~\ref{tab:dataset_stats} summarizes the event and graph level statistics of the Twitter15 dataset after preprocessing.

\section{Method}

\begin{figure}[t]
\centering
\includegraphics[width=0.95\linewidth]{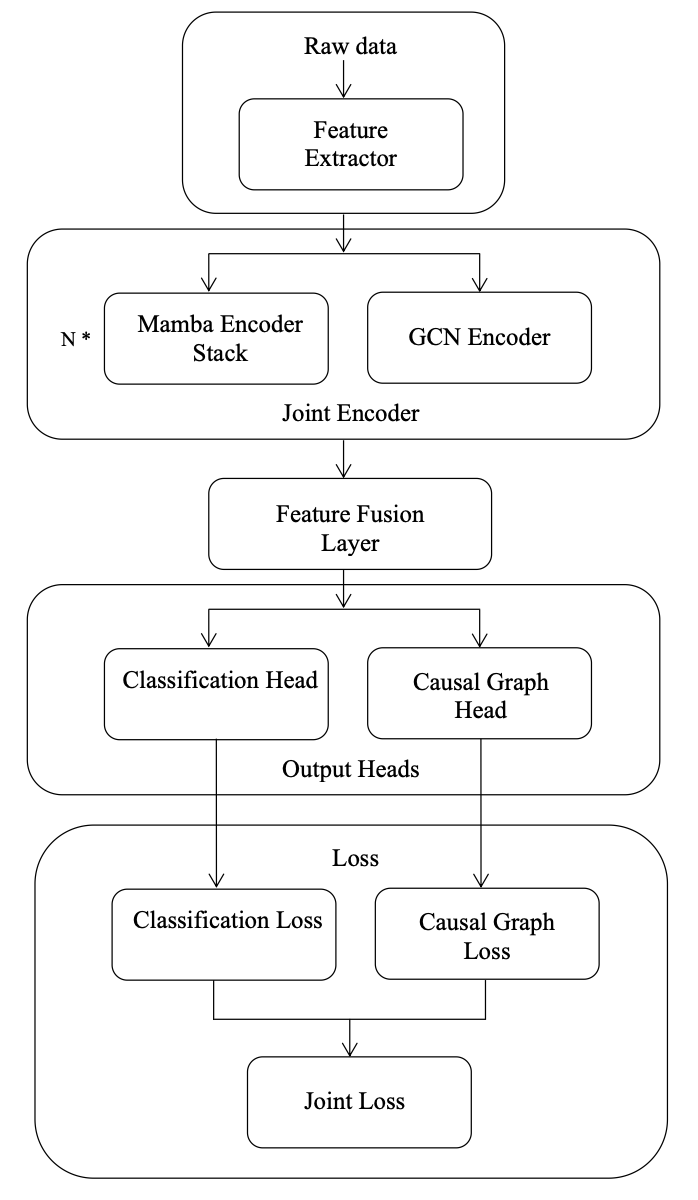}
\caption{
Overview of CausalMamba. Given an input propagation chain, node features are encoded via Mamba and GCN encoders. The fused representation is used for rumor classification and causal graph discovery, optimized jointly via a multi-task loss.
}
\label{fig:architecture}
\end{figure}

We propose \textbf{CausalMamba}, a unified architecture for jointly modeling rumor classification and causal graph discovery in social media propagation chains. The model combines a Mamba-based sequence encoder, a GCN structural encoder, and a differentiable causal learner inspired by NOTEARS. Figure~\ref{fig:architecture} illustrates the full architecture.

\subsection{Problem Formulation}

Each propagation event is modeled as a graph $G = (V, E)$, where nodes $v_i \in V$ represent tweets and edges $E$ encode reply or retweet relationships. Given node features $x_i$, the task is to predict the veracity label $y \in \{\texttt{True}, \texttt{False}, \texttt{Unverified}, \texttt{Non-rumor}\}$, and simultaneously recover a causal adjacency matrix $\hat{A}$ capturing directional influence among tweets.

\subsection{Sequence Encoding via Mamba}

We adopt a two-layer Mamba encoder~\citep{gu2023mamba} to model long-range sequential dependencies. Given batched input $X \in \mathcal{R}^{B \times L \times 833}$ and mask $M \in \{0, 1\}^{B \times L}$, the features are projected to hidden dimension $d = 128$, passed through stacked Mamba blocks, and masked at each layer:
\begin{equation}
H^{\text{seq}} = \text{MambaEncoder}(X, M)
\end{equation}
Each Mamba layer contains a selective state-space block, dropout, and residual normalization.

\subsection{Structural Encoding via GCN}

To capture the graph topology, we apply a single-layer Graph Convolutional Network (GCN)~\citep{kipf_semi-supervised_2017}. The flattened Mamba output is passed to:
\begin{equation}
H^{\text{graph}} = \text{GCN}(H^{\text{seq}}, \texttt{edge\_index})
\end{equation}
Padding nodes are masked to avoid influence propagation.

\subsection{Feature Fusion and Pooling}

The sequence and structure encoders are combined via a residual fusion strategy:
\begin{equation}
H = H^{\text{seq}} + \alpha \cdot H^{\text{graph}}, \quad \alpha = 0.3
\end{equation}
We then apply masked mean pooling over nodes to obtain an event-level vector:
\begin{equation}
z = \frac{1}{\sum_i M_i} \sum_{i=1}^L H_i \cdot M_i
\end{equation}

\subsection{Rumor Classification}

The pooled vector $z$ is fed into a two-layer feed-forward network with ReLU activation and softmax:
\begin{equation}
\hat{y} = \text{softmax}(W_2 \cdot \text{ReLU}(W_1 z + b_1) + b_2)
\end{equation}
We use smoothed cross-entropy loss:
\begin{equation}
\mathcal{L}_{\text{cls}} = (1 - \epsilon) \cdot \text{CE}(y, \hat{y}) + \epsilon \cdot \text{Uniform}, \epsilon = 0.1
\end{equation}

\subsection{Causal Graph Discovery}

To learn causal dependencies, we extract the masked hidden states $H' \in \mathcal{R}^{n \times d}$ and compute:
\begin{equation}
W = \frac{H' H'^\top}{d}
\end{equation}
Then, we define a NOTEARS-style loss:
\begin{align}
\mathcal{L}_{\text{causal}} = &\ \|H' - W H'\|^2 + \lambda_1 \|W\|_1  \nonumber\\
&+ \lambda_2 (\text{tr}(\exp(W \odot W)) - n)
\end{align}
Here, the last term enforces acyclicity~\citep{zheng_dags_2018}. We compute this loss using the first graph in each batch to avoid padding effects.

\subsection{Joint Optimization}

The total loss is a weighted combination of classification and causal terms:
\begin{equation}
\mathcal{L}_{\text{total}} = \mathcal{L}_{\text{cls}} + \lambda \cdot \mathcal{L}_{\text{causal}}, \quad \lambda = 0.1
\end{equation}
We use the AdamW optimizer with learning rate $5 \times 10^{-5}$, weight decay 0.05, gradient clipping (max-norm 1.0), and early stopping based on macro-F1.

\subsection{Causal Intervention Simulation}

At inference time, we extract the learned causal graph $W$, identify top nodes via PageRank, and simulate node removal to visualize causal effects. We compare the original and intervened graphs to evaluate node-level influence on information propagation.

\section{Experiments}

We evaluate CausalMamba on the Twitter15 dataset for 4-way rumor classification and causal graph discovery. This section describes our experimental setup, baseline comparisons, ablation analysis, and qualitative results on causal intervention.

\subsection{Experimental Setup}

We conduct experiments on the Twitter15 dataset~\citep{ma2017detect}, where each propagation event is annotated with one of four labels: \textit{True}, \textit{False}, \textit{Unverified}, or \textit{Non-rumor}. Each event consists of a source tweet followed by a reply/retweet chain that defines the propagation structure.

We split the dataset into 70\% training, 15\% validation, and 15\% test sets. We use Accuracy and Macro-F1 as evaluation metrics to account for class imbalance. Models are trained with batch size 16, hidden dimension 128, dropout 0.2, and early stopping (patience=10) based on validation F1. We use the AdamW optimizer with learning rate $5\text{e}{-5}$ and weight decay 0.05. All experiments are implemented using PyTorch, HuggingFace Transformers, and PyTorch Geometric.

\subsection{Baseline Comparison}

We compare CausalMamba with several representative baselines, including both sequential and structure-aware models. Table~\ref{tab:model_performance} summarizes the performance of each model on the test set.

\paragraph{BiLSTM-CNN} A sequential baseline adapted from~\citet{asghar_exploring_2021}, which encodes tweet sequences using a bidirectional LSTM followed by a CNN over temporal positions.

\paragraph{Transformer} A vanilla transformer encoder applied over tweet sequences. This model replaces RNNs with multi-head attention for long-range modeling, but ignores graph structure.

\paragraph{Mamba} A recent state-space model that replaces attention with a selective input-state update mechanism. Mamba achieves linear time complexity and excels at long-sequence modeling.

\paragraph{Mamba-GCN} This hybrid model augments Mamba with graph structural encodings from GCN. Node embeddings from Mamba are fused with structure-aware outputs, combining sequential and relational signals.

\paragraph{CausalMamba} Extends Mamba-GCN with a causal discovery module based on NOTEARS. In addition to classification, the model jointly learns a latent causal graph, optimized via acyclicity-constrained loss.

\begin{table}[h]
\centering
\begin{tabular}{lcc}
\hline
\textbf{Model} & \textbf{Accuracy} & \textbf{Macro F1} \\
\hline
BiLSTM-CNN    & 0.505 & 0.504 \\
Transformer   & 0.522 & 0.522 \\
Mamba         & 0.594 & 0.590 \\
Mamba-GCN     & 0.643 & 0.643 \\
CausalMamba   & 0.597 & 0.598 \\
\hline
\end{tabular}
\caption{Performance comparison of different models.}
\label{tab:model_performance}
\end{table}

The BiLSTM-CNN baseline achieves moderate performance, which is slightly improved by replacing the encoder with a Transformer. Introducing the Mamba state space model leads to a notable gain of +7\% in both accuracy and F1, demonstrating its ability to model long-range dependencies efficiently. Further incorporating GCN structure encoding yields the strongest results, suggesting that structural signals in propagation graphs are highly informative for veracity prediction.

Interestingly, the full CausalMamba model, which includes an auxiliary causal discovery module — shows a small drop in classification metrics compared to Mamba-GCN. This reflects a common trade-off between predictive accuracy and interpret-ability, as the model now optimizes a joint loss balancing classification and causal structure learning.

\subsection{Layered Ablation Analysis}

Rather than performing isolated ablations, we adopt a layered experimental design where each model incrementally builds on the previous. This design allows us to attribute performance gains to specific architectural components:

\paragraph{BiLSTM-CNN} Serves as a classical RNN-based baseline for rumor detection.

\paragraph{Transformer} Introduces pre-trained contextual embeddings, improving generalization.

\paragraph{Mamba} Replaces the Transformer with a more efficient long-sequence model, improving accuracy by 7\%.

\paragraph{Mamba-GCN} Incorporates propagation structure via GCN, adding another 5\% gain.

\paragraph{CausalMamba} Adds causal supervision, which slightly lowers performance but enables interpretability.
    
This ablation confirms the additive benefits of each component and highlights the robustness of the underlying Mamba encoder.

\subsection{Causal Discovery and Intervention}

To analyze the interpret-ability of our model, we visualize the causal graphs learned by the CausalMamba architecture using a differentiable NOTEARS module ~\citep{zheng_dags_2018}. For each event, we extract the final hidden states, compute a soft adjacency matrix, and simulate counterfactual interventions by removing top-ranked nodes based on PageRank scores.

\begin{figure*}[!htbp]
\centering
\begin{subfigure}[t]{0.48\textwidth}
    \centering
    \includegraphics[width=\linewidth]{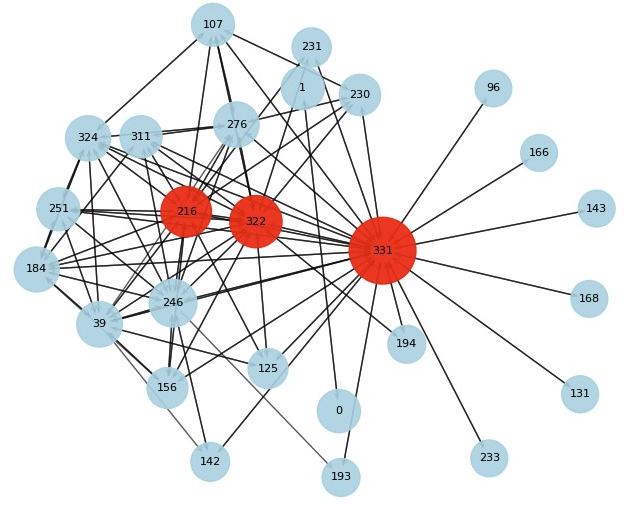}
    \caption{Before intervention.}
    \label{fig:intervention-before}
\end{subfigure}
\hfill
\begin{subfigure}[t]{0.48\textwidth}
    \centering
    \includegraphics[width=\linewidth]{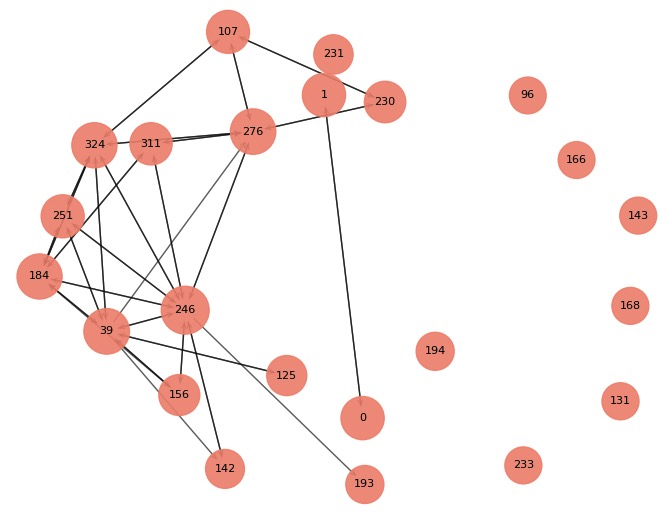}
    \caption{After intervention.}
    \label{fig:intervention-after}
\end{subfigure}
\caption{
Causal graph visualization before and after node intervention. Top-3 nodes (in red) are identified via PageRank. Their removal significantly disrupts information flow, illustrating the model’s interpretability and intervention capability.
}
\label{fig:intervention}
\end{figure*}

Figure~\ref{fig:intervention} shows an example from Event 17, which the model predicts as a rumor with high confidence. The left graph illustrates the learned causal structure before intervention, where nodes \texttt{331}, \texttt{322}, and \texttt{39} exhibit dominant influence over the propagation flow. These nodes are highlighted in red and correspond to the top-3 ranked nodes.

Upon removal of these nodes (right), the causal graph becomes fragmented and disconnected. This demonstrates the potential of our model not only to classify misinformation but also to identify key drivers in its spread—enabling actionable interventions.

This demonstrates the potential of our model not only to classify misinformation, but also to identify key drivers in its spread and enable actionable interventions.

\section{Analysis}

Beyond classification performance, a core motivation of our framework is to provide interpretable insights into rumor propagation by discovering latent causal structures. In this section, we analyze the causal graphs learned by CausalMamba, highlight influential nodes, simulate counterfactual interventions, and explore the interaction between structural and sequential components.

\subsection{Causal Graph Structure and Influential Nodes}

CausalMamba integrates a differentiable causal discovery module that learns a directed acyclic graph (DAG) over tweet nodes within each propagation event. To quantify influence, we compute PageRank scores on the learned adjacency matrix for each event. These scores identify nodes that dominate information flow within the inferred causal structure.

Interestingly, we observe that the top-ranked nodes are not always the source tweet. In many cases, they correspond to intermediate replies that act as central bridges in the reply tree—often accumulating retweets or triggering further discussion. For example, in Event 17 (Figure~\ref{fig:intervention}), nodes \texttt{331}, \texttt{322}, and \texttt{39} exhibit the highest PageRank scores. These nodes occupy structurally central positions and appear early in the temporal chain, suggesting a strong role in downstream propagation.

This finding validates that the causal learner is not simply capturing position or frequency, but encoding nuanced relational influence that aligns with human intuition about rumor spreaders.

\subsection{Counterfactual Intervention Effects}

To assess the utility of the learned causal graphs, we simulate counterfactual interventions by removing the top-k influential nodes and inspecting the resulting graph structure. Figure~\ref{fig:intervention} shows the causal graph of a detected rumor before (left) and after (right) intervention.

Before intervention, the graph forms a well-connected structure with multiple downstream paths. After removing the top-3 nodes identified via PageRank, the graph fragments into disconnected components, indicating a significant disruption in information flow. This qualitative change confirms that the model is not only detecting rumors, but also identifying actionable control points within the propagation graph.

Such intervention simulation is not possible with traditional transformer or GNN-based classifiers alone. The explicit modeling of causal dependencies provides a mechanism for evaluating the effect of potential moderation or node-level suppression in real-world rumor tracking systems.

\subsection{Interplay of Content, Structure, and Causality}

While CausalMamba combines sequential, structural, and causal reasoning modules, their relative contributions vary across events. We find that long and deep propagation chains benefit more from the inclusion of GCN-based structure encoding. In contrast, short or noisy chains rely more heavily on the content modeling capabilities of the Mamba encoder.

The causal loss introduces an auxiliary objective that improves interpretability but may occasionally trade off predictive performance. For example, the causal module encourages sparsity and acyclicity, which can penalize dense graphs or noisy hidden representations. This aligns with our quantitative findings where CausalMamba exhibits slightly lower classification scores than Mamba-GCN, but gains significantly in structure interpretability and intervention capacity.

\subsection{Future Work}

This work introduces CausalMamba as a joint sequence-structure-causality framework for rumor detection. While initial results are promising, several directions remain for further development and validation.

First, our current evaluation is restricted to the Twitter15 dataset. Extending to other rumor detection corpora such as Twitter16, PHEME, and Weibo will help assess the generalization of the model under different linguistic, structural, and temporal distributions. Notably, these datasets may vary in average chain length, branching patterns, and label balance, offering a more rigorous testbed for causal modeling.

Second, while CausalMamba achieves competitive classification performance, its accuracy slightly lags behind the Mamba-GCN baseline due to the introduction of causal constraints. In future work, we plan to explore dynamic loss weighting strategies, where the relative weight of the causal loss ($\lambda$) is adaptively adjusted based on event complexity or model confidence. Additionally, a multi-objective optimization framework could be introduced to better balance predictive accuracy and causal interpret-ability.

Third, our current causal discovery module is used primarily for identifying influential nodes via PageRank and simulating interventions. Future research could enhance this component by incorporating causal saliency methods to more directly connect causal structure with classification decisions. Furthermore, introducing weak supervision signals—such as annotated rumor sources or partial reply trees—could guide the learning of more semantically meaningful causal graphs.

Finally, we envision integrating the CausalMamba framework into real-world rumor moderation pipelines. The learned causal graphs could support user-facing tools for rumor tracking and intervention planning, such as visualization dashboards that highlight high-influence nodes and project the expected impact of content moderation actions. Such integration would not only validate the practical value of our model but also contribute toward responsible and interpretable AI systems in high-stakes information environments.

\section{Conclusion}

We present CausalMamba, a unified framework that jointly models temporal dynamics, structural dependencies, and causal influence for social media rumor detection. By integrating a Mamba-based sequence encoder, a GCN-based structure learner, and a differentiable NOTEARS module, our model captures the multifaceted nature of rumor propagation. Our approach not only improves classification accuracy over BiLSTM and Transformer baselines, but also offers model interpretability through causal graph discovery. The joint training scheme encourages sparse, acyclic influence structures aligned with the propagation chain, while retaining competitive predictive power. The learned causal graphs enable targeted intervention: removing top-ranked nodes identified by PageRank significantly alters the structure and flow of misinformation, illustrating the model’s utility in real-world mitigation efforts. We believe this work provides a foundation for explainable rumor detection systems that combine predictive accuracy with actionable insights. It bridges the gap between neural sequence modeling and causal reasoning in dynamic social contexts. Future work may extend this framework to multimodal misinformation, incorporate more expressive causal learners, and explore deployment strategies under adversarial or real-time settings.

\bibliography{custom}

@misc{dziri2023faith,
	title = {Faith and {Fate}: {Limits} of {Transformers} on {Compositionality}},
	shorttitle = {Faith and {Fate}},
	url = {http://arxiv.org/abs/2305.18654},
	doi = {10.48550/arXiv.2305.18654},
	urldate = {2025-05-12},
	publisher = {arXiv},
	author = {Dziri, Nouha and Lu, Ximing and Sclar, Melanie and Li, Xiang Lorraine and Jiang, Liwei and Lin, Bill Yuchen and West, Peter and Bhagavatula, Chandra and Bras, Ronan Le and Hwang, Jena D. and Sanyal, Soumya and Welleck, Sean and Ren, Xiang and Ettinger, Allyson and Harchaoui, Zaid and Choi, Yejin},
	month = oct,
	year = {2023},
	note = {arXiv:2305.18654 [cs]},
}

@misc{gu2023mamba,
	title = {Mamba: {Linear}-{Time} {Sequence} {Modeling} with {Selective} {State} {Spaces}},
	shorttitle = {Mamba},
	url = {http://arxiv.org/abs/2312.00752},
	doi = {10.48550/arXiv.2312.00752},
	language = {en},
	urldate = {2025-04-01},
	publisher = {arXiv},
	author = {Gu, Albert and Dao, Tri},
	month = may,
	year = {2024},
	note = {arXiv:2312.00752 [cs]},
}

@misc{peng2024limitations,
	title = {On {Limitations} of the {Transformer} {Architecture}},
	url = {http://arxiv.org/abs/2402.08164},
	doi = {10.48550/arXiv.2402.08164},
	urldate = {2025-05-12},
	publisher = {arXiv},
	author = {Peng, Binghui and Narayanan, Srini and Papadimitriou, Christos},
	month = feb,
	year = {2024},
	note = {arXiv:2402.08164 [stat]},
}

@inproceedings{gu2022s4,
  title={Efficiently Modeling Long Sequences with Structured State Spaces},
  author={Gu, Albert and Goel, Karan and Re, Christopher},
  booktitle={International Conference on Learning Representations (ICLR)},
  year={2022},
  url={https://github.com/HazyResearch/state-spaces}
}

@inproceedings{gupta2022dss,
  title={Diagonal State Spaces are as Effective as Structured State Spaces},
  author={Gupta, Ankit and Gu, Albert and Berant, Jonathan},
  booktitle={Neural Information Processing Systems (NeurIPS)},
  year={2022},
  url={https://github.com/ag1988/dss}
}

@inproceedings{lu2023s5,
  title={Structured State Space Models for In-Context Reinforcement Learning},
  author={Lu, Chris and Schroecker, Yannick and Gu, Albert and Parisotto, Emilio and Foerster, Jakob and Singh, Satinder and Behbahani, Feryal},
  booktitle={Neural Information Processing Systems (NeurIPS)},
  year={2023},
  url={https://github.com/luchris429/s5rl}
}

@article{wu_misinformation_2019,
	title = {Misinformation in {Social} {Media}: {Definition}, {Manipulation}, and {Detection}},
	volume = {21},
	issn = {1931-0145},
	shorttitle = {Misinformation in {Social} {Media}},
	url = {https://dl.acm.org/doi/10.1145/3373464.3373475},
	doi = {10.1145/3373464.3373475},
	number = {2},
	urldate = {2025-04-01},
	journal = {SIGKDD Explor. Newsl.},
	author = {Wu, Liang and Morstatter, Fred and Carley, Kathleen M. and Liu, Huan},
	month = nov,
	year = {2019},
	pages = {80--90},
}

@inproceedings{cheng_causal_2021,
	address = {New York, NY, USA},
	series = {{KDD} '21},
	title = {Causal {Understanding} of {Fake} {News} {Dissemination} on {Social} {Media}},
	isbn = {978-1-4503-8332-5},
	url = {https://dl.acm.org/doi/10.1145/3447548.3467321},
	doi = {10.1145/3447548.3467321},
	urldate = {2025-04-01},
	booktitle = {Proceedings of the 27th {ACM} {SIGKDD} {Conference} on {Knowledge} {Discovery} \& {Data} {Mining}},
	publisher = {Association for Computing Machinery},
	author = {Cheng, Lu and Guo, Ruocheng and Shu, Kai and Liu, Huan},
	month = aug,
	year = {2021},
	pages = {148--157},
}

@article{asghar_exploring_2021,
	title = {Exploring deep neural networks for rumor detection},
	volume = {12},
	issn = {1868-5145},
	url = {https://doi.org/10.1007/s12652-019-01527-4},
	doi = {10.1007/s12652-019-01527-4},
	language = {en},
	number = {4},
	urldate = {2025-04-01},
	journal = {Journal of Ambient Intelligence and Humanized Computing},
	author = {Asghar, Muhammad Zubair and Habib, Ammara and Habib, Anam and Khan, Adil and Ali, Rehman and Khattak, Asad},
	month = apr,
	year = {2021},
	pages = {4315--4333},
}

@incollection{li_causal_2023,
	address = {Cham},
	title = {Causal {Inference} and {Natural} {Language} {Processing}},
	isbn = {978-3-031-35050-4 978-3-031-35051-1},
	url = {https://link.springer.com/10.1007/978-3-031-35051-1_9},
	language = {en},
	urldate = {2025-04-02},
	booktitle = {Machine {Learning} for {Causal} {Inference}},
	publisher = {Springer International Publishing},
	author = {Chen, Wenqing and Chu, Zhixuan},
	editor = {Li, Sheng and Chu, Zhixuan},
	year = {2023},
	doi = {10.1007/978-3-031-35051-1_9},
	pages = {189--206},
}

@article{feder_causal_2022,
	title = {Causal inference in natural language processing: {Estimation}, prediction, interpretation and beyond},
	volume = {10},
	shorttitle = {Causal inference in natural language processing},
	url = {https://direct.mit.edu/tacl/article-abstract/doi/10.1162/tacl_a_00511/113490},
	urldate = {2025-04-02},
	journal = {Transactions of the Association for Computational Linguistics},
	author = {Feder, Amir and Keith, Katherine A. and Manzoor, Emaad and Pryzant, Reid and Sridhar, Dhanya and Wood-Doughty, Zach and Eisenstein, Jacob and Grimmer, Justin and Reichart, Roi and Roberts, Margaret E.},
	year = {2022},
	note = {Publisher: MIT Press One Broadway, 12th Floor, Cambridge, Massachusetts 02142, USA …},
	pages = {1138--1158},
}

@misc{zheng_dags_2018,
	title = {{DAGs} with {NO} {TEARS}: {Continuous} {Optimization} for {Structure} {Learning}},
	shorttitle = {{DAGs} with {NO} {TEARS}},
	url = {http://arxiv.org/abs/1803.01422},
	doi = {10.48550/arXiv.1803.01422},
	urldate = {2025-04-02},
	publisher = {arXiv},
	author = {Zheng, Xun and Aragam, Bryon and Ravikumar, Pradeep and Xing, Eric P.},
	month = nov,
	year = {2018},
	note = {arXiv:1803.01422 [stat]},
}

@misc{yu_dag-gnn_2019,
	title = {{DAG}-{GNN}: {DAG} {Structure} {Learning} with {Graph} {Neural} {Networks}},
	shorttitle = {{DAG}-{GNN}},
	url = {http://arxiv.org/abs/1904.10098},
	doi = {10.48550/arXiv.1904.10098},
	urldate = {2025-04-02},
	publisher = {arXiv},
	author = {Yu, Yue and Chen, Jie and Gao, Tian and Yu, Mo},
	month = apr,
	year = {2019},
	note = {arXiv:1904.10098 [cs]},
}

@article{hu2022causal,
	title = {Causal {Inference} for {Leveraging} {Image}-{Text} {Matching} {Bias} in {Multi}-{Modal} {Fake} {News} {Detection}},
	volume = {35},
	issn = {1558-2191},
	url = {https://ieeexplore.ieee.org/abstract/document/9996587?casa_token=0g4o1RdC6C4AAAAA:upXS4sr2Fuanse-9jegNeRG9B-CjiV7E4zCewEuh_mvdWhi09u3PyUiulMykPYWxD-zakSmL},
	doi = {10.1109/TKDE.2022.3231338},
	number = {11},
	urldate = {2025-04-02},
	journal = {IEEE Transactions on Knowledge and Data Engineering},
	author = {Hu, Linmei and Chen, Ziwei and Zhao, Ziwang and Yin, Jianhua and Nie, Liqiang},
	month = nov,
	year = {2023},
	note = {Conference Name: IEEE Transactions on Knowledge and Data Engineering},
	pages = {11141--11152},
}

@inproceedings{ma2017detect,
    title = "Detect Rumors in Microblog Posts Using Propagation Structure via Kernel Learning",
    author = "Ma, Jing  and
      Gao, Wei  and
      Wong, Kam-Fai",
    editor = "Barzilay, Regina  and
      Kan, Min-Yen",
    booktitle = "Proceedings of the 55th Annual Meeting of the Association for Computational Linguistics (Volume 1: Long Papers)",
    month = jul,
    year = "2017",
    address = "Vancouver, Canada",
    publisher = "Association for Computational Linguistics",
    url = "https://aclanthology.org/P17-1066/",
    doi = "10.18653/v1/P17-1066",
    pages = "708--717"
}

@article{bian_rumor_2020,
	title = {Rumor {Detection} on {Social} {Media} with {Bi}-{Directional} {Graph} {Convolutional} {Networks}},
	volume = {34},
	copyright = {Copyright (c) 2020 Association for the Advancement of Artificial Intelligence},
	issn = {2374-3468},
	url = {https://ojs.aaai.org/index.php/AAAI/article/view/5393},
	doi = {10.1609/aaai.v34i01.5393},
	language = {en},
	number = {01},
	urldate = {2025-05-11},
	journal = {Proceedings of the AAAI Conference on Artificial Intelligence},
	author = {Bian, Tian and Xiao, Xi and Xu, Tingyang and Zhao, Peilin and Huang, Wenbing and Rong, Yu and Huang, Junzhou},
	month = apr,
	year = {2020},
	note = {Number: 01},
	pages = {549--556},
}

@inproceedings{dong2019multiple,
	address = {New York, NY, USA},
	series = {{CIKM} '19},
	title = {Multiple {Rumor} {Source} {Detection} with {Graph} {Convolutional} {Networks}},
	isbn = {978-1-4503-6976-3},
	url = {https://dl.acm.org/doi/10.1145/3357384.3357994},
	doi = {10.1145/3357384.3357994},
	urldate = {2025-05-11},
	booktitle = {Proceedings of the 28th {ACM} {International} {Conference} on {Information} and {Knowledge} {Management}},
	publisher = {Association for Computing Machinery},
	author = {Dong, Ming and Zheng, Bolong and Quoc Viet Hung, Nguyen and Su, Han and Li, Guohui},
	month = nov,
	year = {2019},
	pages = {569--578},
}

@misc{kipf_semi-supervised_2017,
	title = {Semi-{Supervised} {Classification} with {Graph} {Convolutional} {Networks}},
	url = {http://arxiv.org/abs/1609.02907},
	doi = {10.48550/arXiv.1609.02907},
	urldate = {2025-05-12},
	publisher = {arXiv},
	author = {Kipf, Thomas N. and Welling, Max},
	month = feb,
	year = {2017},
	note = {arXiv:1609.02907 [cs]},
}

\appendix

\section{Ethics Statement}
\label{app:ethics}

This work utilizes the Twitter15 dataset, which is publicly available and widely used in rumor detection research. All tweet content is accessed via tweet IDs in compliance with Twitter’s terms of use. No personally identifiable information is collected, stored, or analyzed. Our analysis focuses solely on aggregated propagation patterns and does not make inferences about individual users. This study involves no human subjects and aligns with ethical principles of data minimization and responsible AI.

\section{Experimental Settings}
\label{app:implementation}

\subsection{Experimental Environment}

Our experiments were conducted on the Pittsburgh Supercomputing Center (PSC) using an NVIDIA H100 GPU. We used a fixed random seed (\texttt{set\_seed(42)}) for reproducibility. Training time for \textsc{CausalMamba} on Twitter15 was approximately 10 minutes.

\noindent\textbf{Software Stack:}

\begin{table}[h]
\centering
\begin{tabular}{l|l}
\hline
\textbf{Component} & \textbf{Version} \\
\hline
Python & 3.12.7 \\
PyTorch & 2.2.0 \\
Transformers & 4.38.0 \\
Mamba library & \texttt{mamba-ssm} (official) \\
Environment manager & Anaconda \\
\hline
\end{tabular}
\caption{Software and library versions used in our experiments.}
\label{tab:env}
\end{table}

\noindent\textbf{Code and Data.} Available at: \url{https://github.com/XiaotongZhan/info259-final-project}

\subsection{Hyper-parameters}

Table \ref{tab:hyperparams} lists the hyperparameter ranges used in our experiments.

\begin{table}[!htbp]
\centering
\begin{tabular}{l|p{0.5\linewidth}}
\hline
\textbf{Parameter} & \textbf{Value} \\
\hline
Input dimension & 833 (BERT+Time+User) \\
Hidden dimension & 128 \\
Mamba layers & 2 \\
Batch size & 16 \\
Dropout & 0.2 \\
Optimizer & AdamW \\
Learning rate & 5e-5 \\
Weight decay & 5e-2 \\
Label smoothing & 0.1 \\
Gradient clip norm & 1.0 \\
Causal loss weight $\lambda$ & 1.0 \\
Early stopping patience & 10 epochs \\
Pooling strategy & Mean pooling \\
Fusion ratio & 1.0(Mamba)+0.3 (GCN) \\
\hline
\end{tabular}
\caption{Hyperparameter settings for \textsc{CausalMamba}.}
\label{tab:hyperparams}
\end{table}

\end{document}